\documentclass[11pt, a4paper]{article}

\usepackage[english]{babel}
\usepackage{verbatim}
\usepackage{epstopdf}
\usepackage{amsfonts} 
\usepackage{amsthm}
\usepackage{amsmath}
\usepackage{amssymb}
\usepackage{color}
\usepackage{textcomp} 
\usepackage{rotating}
\usepackage{graphicx}
\usepackage{caption}
\usepackage{subcaption}
\usepackage{paralist} 
\usepackage[pdftex, pdfborder={0 0 0 0}]{hyperref} 
\usepackage{booktabs}
\usepackage{bbm}
\usepackage{algorithm,algorithmic}

\usepackage[font={small,it}]{caption}
\usepackage{float}
\usepackage{setspace}
\usepackage{adjustbox}
\usepackage{booktabs}
\usepackage{longtable}
\usepackage[margin=1in]{geometry}

\usepackage{epsfig}
\usepackage{ulem}
\usepackage{natbib}

\newcommand{\bls}[1]{\renewcommand{\baselinestretch}{#1}\footnotesize\normalsize}

\newenvironment{changemargin}[2]{%
  \begin{list}{}{%
    \setlength{\topsep}{0pt}%
    \setlength{\leftmargin}{#1}%
    \setlength{\rightmargin}{#2}%
    \setlength{\listparindent}{\parindent}%
    \setlength{\itemindent}{\parindent}%
    \setlength{\parsep}{\parskip}%
  }%
  \item[]}{\end{list}}

\begin{document}
\title{Ensemble of Precision-Recall Curve (PRC) Classification Trees with Autoencoders
\vspace{-4mm}}
\author{
{\large Jiaju  \ Miao}$^{a\,}$\footnote{Corresponding author at: Department of Applied Mathematics and Statistics, Stony Brook University, Stony Brook, New York, United States. E-mail address: \texttt{miaojiaju@gmail.com}.}
 \hspace*{2mm}
{\large Wei \ Zhu}$^{a\,}$
\\ \vspace{-2mm}
$^{a}$\textit{\small Department of Applied Mathematics and Statistics, Stony Brook University, United States}\\
}
\date{\today}
\maketitle \thispagestyle{empty}
\vspace*{-12mm}
\begin{abstract}
\begin{changemargin}{-5mm}{-5mm}
\bls{1.2}
Anomaly detection underpins critical applications—from network security and intrusion detection to fraud prevention—where recognizing aberrant patterns rapidly is indispensable. Progress in this area is routinely impeded by two obstacles: extreme class imbalance and the curse of dimensionality. To combat the former, we previously introduced Precision–Recall Curve (PRC) classification trees and their ensemble extension, the PRC Random Forest (PRC-RF). Building on that foundation, we now propose a hybrid framework that integrates PRC-RF with autoencoders—unsupervised machine learning methods that learn compact latent representations—to confront both challenges simultaneously. Extensive experiments across diverse benchmark datasets demonstrate that the resulting Autoencoder-PRC-RF model achieves superior accuracy, scalability, and interpretability relative to prior methods, affirming its potential for high-stakes anomaly-detection tasks.

\end{changemargin}
\end{abstract}
\vspace{-2mm}
\pagebreak

\section{Introduction}

The joint use of autoencoders and random forests offers a powerful framework for tackling complex data-analysis tasks, especially when the data are both large and high-dimensional. Autoencoders—unsupervised neural networks—excel at uncovering intricate, non-linear relationships and learning compact latent representations; as shown by \citet{Hinton:06}, deep autoencoder architectures can reduce dimensionality with remarkable fidelity. Random forests, by contrast, are ensemble learners suited to both classification and regression and are particularly resilient in noisy, high-dimensional settings.

The combination of these two methods involves using an autoencoder to pre-process the data by reducing its dimensionality and extracting useful features (\citet{Vincent:08}), which are then used as inputs to a random forest model. This can help to improve the accuracy and generalizability of the random forest model, while also providing insight into the underlying structure of the data.

Beyond scalability to “big” data, the autoencoder–random-forest combination is notably robust: it maintains performance in the face of noise, missing entries, and only weakly labeled samples—conditions under which many traditional methods falter. Consequently, the approach has gained traction across finance, healthcare, natural-language processing, and other data-rich domains.

Yet dimensionality is only one of the hurdles that modern classifiers face. Severe class imbalance is another, arising when the minority class—often the outcome of greatest interest (e.g., fraud cases, diseased patients)—is dwarfed by the majority class. To address this, we previously introduced Precision–Recall Curve (PRC) classification trees and their ensemble extension, the PRC Random Forest (PRC-RF) \citep{Miao:22}. In the present study, we integrate deep autoencoders with PRC-RF, yielding the \emph{Autoencoder–PRC-RF} model. This unified architecture simultaneously mitigates the curse of dimensionality and the skewed-class problem, delivering a robust solution for anomaly-detection tasks.

The remainder of the paper is organized as follows. Section~\ref{sec:Methodology} reviews the PRC-RF classifier and summarizes autoencoders for anomaly detection. Section~\ref{sec:ensemble} details our strategy for integrating autoencoder features with PRC tree-based methods. Section~\ref{sec:casestudy} reports an empirical evaluation across diverse, real-world scenarios. Finally, Section~\ref{sec:conclusion} concludes the paper and outlines directions for future research.

\section{Methodology}\label{sec:Methodology}

We begin by providing a comprehensive overview of the PRC classification trees developed in our prior work \citet{Miao:22}, along with neural network-based autoencoders, with a particular emphasis on their applications in anomaly detection. We revisit the theoretical underpinnings of the PRC framework and highlight its distinctions from traditional classification trees, especially in handling imbalanced datasets and skewed class distributions. Furthermore, we extend the discussion to the role of autoencoders—a widely adopted deep learning technique—in the anomaly detection paradigm. As noted in \citet{Pang:21}, recent advances have demonstrated the efficacy of deep learning models, including autoencoders, in various anomaly detection tasks.

\subsection{PRC Random Forest Classifier}
We begin with the central criterion underpinning the PRC classification tree of \citet{miao:thesis}: the area under the precision–recall curve (AUPRC). AUPRC is a standard metric for binary classification, especially when class distributions are highly imbalanced. A precision–recall (PR) curve traces, across all decision thresholds, the trade-off between \emph{precision}—the fraction of predicted positives that are correct—and \emph{recall}—the fraction of actual positives that are recovered. Integrating this curve yields the AUPRC, a single scalar that summarizes how well a classifier retrieves positive instances both accurately and comprehensively. Higher AUPRC values therefore reflect a more favorable balance between precision and recall, making the metric indispensable for ranking models in the imbalanced-data settings common to real-world applications. For AUPRC, the baseline value is the ratio of positive cases in the distribution expressed as
\begin{equation}
Baseline \, Value =\frac{positive\,cases}{positive\,cases+negative\, cases} .
\end{equation}

Algorithm \ref{auprc} below shows the process of calculating the AUPRC and the computation in the case of it less than the baseline value. For the estimation of AUPRC, we also applied the trapezoidal approach: 
\begin{equation}
AUPRC_{trapezoid}=\frac{1}{2}\cdot\sum_{i=1}^{n}(r_{i+1}-r_{i})\cdot(p_{i+1}+p_{i})
\end{equation}
where r is the function of recall and p is of precision, the trajectory is partitioned in n-1 sections. See \citet{Boyd:13} for more details.
\begin{algorithm}[H]
\begin{algorithmic}[1]
\REQUIRE Training data $(x_1,y_1 )$,\ldots,$(x_n,y_n )$, Feature set $f_i \in F$, target vector $y_i \in \{-1,+1\}$. 
\ENSURE Area under precision-recall curve (AUPRC) of feature $f_i$.
\STATE \textbf{sort} data $(X[f_i ], Y)$ by feature
\STATE \textbf{set} $uniq\_ values_{f_i} \leftarrow sort (unique(f_i))$
\STATE \textbf{set} $total\_ positives \leftarrow length (which(Y==1))$
\STATE \textbf{set} $total\_ negatives \leftarrow length (which(Y==-1))$
\STATE \textbf{set} $PRC_{baseline} \leftarrow \frac{total\_ positives}{total\_ positives+total\_ negatives}$
\STATE \textbf{set} $recall\_ array \leftarrow rep(0,length(uniq\_ values_{f_i})$
\STATE \textbf{set} $precision\_ array \leftarrow rep(0,length(uniq\_ values_{f_i})$
\STATE \textbf{set} $AUPRC_{f_i} \leftarrow 0$
\FOR{$j=1$ to $length(uniq\_ values_{f_i})$}
\STATE $indice \leftarrow  which(X[f_i]\leq uniq\_ values_{f_i}[j])$
\STATE $recall\_ array \leftarrow \frac{length(which((X[f_i ],Y)[indice,"Y"]==1))}{total\_ positives}$
\STATE $precision\_ array \leftarrow \frac{length(which((X[f_i ],Y)[indice,"Y"]==1))}{length(indice)}$
\IF{$precision\_ array[j]< PRC_{baseline}$}
\STATE $recall\_ array[j] \leftarrow 1-recall\_ array[j]$
\STATE $precision\_ array[j] \leftarrow \frac{total\_ positives-length(which((X[f_i],Y)[indice,"Y"]==1))}{nrow(X[f_i],Y  )-length(indice)}$
\ENDIF
\IF{$j==1$}
\STATE  $AUPRC_{f_i} \leftarrow \frac{recall\_ array[j]\cdot(1+precision\_ array[j])}{2}$
\ELSE \STATE $AUPRC_{f_i} \leftarrow AUPRC_{f_i}+$\STATE$\frac{(recall\_ array[j]-recall\_ array[j-1])\cdot(precision\_ array[j]+precision\_ array[j-1])}{2}$
\ENDIF
\ENDFOR
\RETURN $AUPRC_{f_i}$
\STATE \textbf{end}
\end{algorithmic}
\caption{AUPRC\_calculation}
\label{auprc}
\end{algorithm}

Feature selection refers to the process of determining which variables to include in a predictive model. In tree-based models, the choice of input features significantly impacts model performance. For the PRC classification tree, feature selection is governed by Algorithm \ref{Feature_selection}. It lets AUPRC decide which feature variable is useful to include in every split of tree node. Typically, the feature that yields the largest AUPRC value should be selected. In Algorithm \ref{Feature_selection}, the function “AUPRC\_calculation” is implemented by Algorithm \ref{auprc}. The below pseudo code illustrates the process of generating the suitable feature with the largest AUPRC value.
\begin{algorithm}[H]
\begin{algorithmic}[1]
\REQUIRE Training data $(x_1,y_1 )$,\ldots,$(x_n,y_n )$, Feature set $f_i \in F$, target vector $y_i \in \{-1,+1\}$.
\ENSURE Feature $f_i$ with the largest AUPRC.
\\ \textbf{Require}: AUPRC\_calculation. 
\STATE \textbf{set} $Max_f  \leftarrow NULL$
\STATE \textbf{set} $Max_{AUPRC} \leftarrow 0$
\FOR{each feature $f_i \in F$}
\STATE $Temp_{AUPRC} \leftarrow AUPRC\_calculation(X[f_i],Y  )$
\IF{$Temp_{AUPRC}> Max_{AUPRC}$}
\STATE $Max_f \leftarrow f_i$
\STATE $Max_{AUPRC} \leftarrow Temp_{AUPRC}$
\ENDIF
\ENDFOR
\RETURN $Max_f, Max_{AUPRC}$
\STATE \textbf{end}
\end{algorithmic}
\caption{Feature\_Selection\_by\_AUPRC}
\label{Feature_selection}
\end{algorithm}

Determining the optimal threshold for each split is a crucial step in constructing the PRC Tree. This is accomplished by selecting the threshold that maximizes the F1-score for the chosen feature. The F1-score, defined as the harmonic mean of precision and recall, serves as a balanced measure of a model’s accuracy in the context of positive class identification.\

Importantly, the F1-score does not consider true negatives, making it particularly well-suited for imbalanced classification tasks where the minority class is of primary interest. Algorithm~\ref{Threshold} outlines the procedure for threshold selection. The main objective is to identify the cutoff value that yields the highest F1-score, based on the set of potential thresholds generated for the selected feature by Algorithm~\ref{Feature_selection}.
\begin{algorithm}[H]
\begin{algorithmic}[1]
\REQUIRE PRC matrix ($recall\_ array,precison\_ array$) of selected feature$f_i$, where $f_i \in F$; Threshold set $\Theta$.
\ENSURE Selected splitting threshold $\theta$.
\\ \textbf{Require}: function $HarmonicMean$. 
\STATE \textbf{set} $Max_{F1}  \leftarrow 0$
\STATE \textbf{set} $Max_{\theta}  \leftarrow NULL$
\STATE \textbf{set} $uniq\_ split_{f_j} \leftarrow sort(unique(f_i))$
\FOR{{$j=1$ to $length(uniq\_ split_{f_j})$}}
\STATE $Temp_{F1} \leftarrow HarmonicMean(recall\_ array[j],precison\_ array[j])$
\IF{$Temp_{F1}> Max_{F1}$}
\STATE $Max_{F1} \leftarrow Temp_{F1}$
\STATE $Max_{\theta} \leftarrow uniq\_ split_{f_j}[j]$
\ENDIF
\ENDFOR
\RETURN $Max_{\theta}$
\STATE \textbf{end}
\end{algorithmic}
\caption{Threshold\_Selection\_by\_F1-score}
\label{Threshold}
\end{algorithm}

         The following Algorithm \ref{tree} is employed to generate the PRC Tree. The previous Algorithm \ref{Feature_selection} and \ref{Threshold} complete a node split process in each stage. Instead of measuring by traditional Gini impurity or information gain, the features and threshold are selected by AUPRC and F1-score respectively. Algorithm  \ref{tree} “PRC\_Tree” integrates the previous algorithms to generate each split node until the corresponding stopping criteria is met. Each node has scores for different classes, measuring the percentage of each class in it. It is called nodescore in the algorithm. The corresponding nodelabel is the majority class of this node. By doing so, it could be easy to achieve the majority target label in the Terminal node. Provided below is the pseudo code to build PRC classification tree recursively. The prediction of a PRC Tree, $\mathcal{T}$, with $K$ terminal nodes and depth $L$, can  be written as
\begin{equation}
g(x_i;\hat{y},K,L)= \sum_{i=1}^{K}\hat{y_{k}}\mathbbm{1}_{\{x_i\in C_k(L)\}},
\end{equation} 
where $C_k(L)$ is one of the $K$ partitions of the data. Each partition is a product of up to $L$ indicator function of the features which are selected by algorithm \ref{Feature_selection}.
\begin{algorithm}[H]
\begin{algorithmic}[1]
\REQUIRE Training data $(x_1,y_1 )$,\ldots,$(x_n,y_n )$, Feature set $f_i \in F$, target vector $y_i \in \{-1,+1\}$; stopping criterion (maximum tree depth, minimum leaf size); $N_f$, the number of features used in each split.
\ENSURE PRC tree.
\\ \textbf{Require}: Feature\_Selection\_by\_AUPRC and Threshold\_Selection\_by\_F1-score
\STATE \textbf{Do} the nodescore and nodelabel for the root node.
\IF{the stopping criterion is met}
\STATE \textbf{return} PRC tree
\ELSE 
\STATE sample $N_f$ features from the feature set $F$
\STATE \textbf{set} the selected features as $F^{'}$
\STATE $(Max_f, Max_{AUPRC}) \leftarrow Feature\_Selection\_by\_AUPRC(X,Y,F^{'})$
\STATE $Max_{\theta} \leftarrow Threshold\_Selection\_by\_F1-score(Max_f, Max_{AUPRC})$
\STATE $\{(X,Y)_{left},(X,Y)_{right}\} \leftarrow Node\_Split(Max_f, Max_{\theta})$
\STATE apply the function $PRC\_Tree\,(maximum tree depth\leftarrow maximum tree depth-1)$ to the subsets $\{(X,Y)_{left},(X,Y)_{right}\}$ recursively until resulting nodes are pure or the stopping criteria is met 
\ENDIF
\RETURN PRC tree
\STATE \textbf{end}
\end{algorithmic}
\caption{PRC\_Tree}
\label{tree}
\end{algorithm}

Random forests are an ensemble learning technique for classification that operate by constructing a collection of decision trees and aggregating their predictions. To build a PRC random forest, we use the PRC tree as the base learner within this ensemble framework. The resulting PRC random forest retains strong predictive performance while also offering reliable estimates of feature importance. Algorithm \ref{rf} “PRC\_Random\_Forest” states how to build the forests based on the PRC tree. The parameter $N_t$ is used to decide the number of trees to form the “forest”. As discussed before, the number of features $N_f$ for each node split is randomly chosen from the feature set F. It can decrease the prediction error of the model by doing so. Below is the procedure of our “forests”.

\begin{algorithm}[H]
\begin{algorithmic}[1]
\REQUIRE Training data $(x_1,y_1 )$,\ldots,$(x_n,y_n )$, Feature set $f_i \in F$, target vector $y_i \in \{-1,+1\}$; Number of trees $N_t$; Number of features for each node split $N_f$.
\ENSURE PRC random forest $\Re$.
\\ \textbf{Require}: PRC\_Tree
\STATE \textbf{Set} $\Re \leftarrow NULL$.
\FOR{{$j=1$ to $N_t$}}
\STATE generate bootstrap sample $(X,Y)^j$
\STATE for each node split, generate $F^{'}$ by randomly choosing $N_f$ features from $F$
\STATE $prc\_tree_j \leftarrow PRC\_Tree((X,Y)^j,F^{'})$
\STATE append $prc\_tree_j\, to\, \Re$
\ENDFOR
\RETURN PRC random forest $\Re$
\STATE \textbf{end}
\end{algorithmic}
\caption{PRC\_Random\_Forest}
\label{rf}
\end{algorithm}

\subsection{Autoencoders for Anomaly Detection}

In recent years, the surge of data‐driven systems and the increasing interconnectedness of diverse domains have underscored the urgent need for effective anomaly‐detection techniques. Anomalies—also called outliers, deviations, or irregularities—often signal critical events, security breaches, or early warnings of system failure. Detecting such events in real time can avert substantial losses and mitigate risk.

Deep learning is a rapidly evolving discipline, with new architectures and training paradigms emerging continuously; see \citet{Good:16} and references therein for a concise overview of recent advances. Among these methods, autoencoders—unsupervised neural networks originally devised for dimensionality reduction—have received considerable attention for their capacity to learn rich representations from high‐dimensional data. By capturing latent structure and recurring patterns, autoencoders have proven well suited to anomaly‐detection tasks. Seminal and subsequent contributions include \citet{Hau:17} and \citet{Munir:19}, among many others.

Generative Adversarial Networks (GANs) extend this toolkit by learning an implicit model of the data distribution, enabling the generation of realistic samples and the identification of anomalies as departures from the learned norm; see, e.g., \citet{deeck:18}, \citet{Akay:18}, \citet{Kaplan:20}, \citet{ze:18}, \citet{Sabo:18}, \citet{schl:17}, and \citet{Akçay:19}. Crucially, GANs (and autoencoders alike) thrive in unsupervised or weakly supervised settings where labeled anomalies are scarce or unavailable.

The core objective of an autoencoder is to reproduce its input as faithfully as possible through an encode–decode cycle. Training proceeds by minimizing the discrepancy—often the mean‐squared or cross‐entropy loss—between the input and its reconstruction. After learning the manifold of normal data, the autoencoder offers a baseline against which new instances are gauged: anomalous samples cannot be reconstructed accurately and thus incur elevated reconstruction errors. Modern anomaly‐detection pipelines exploit these errors, or derived abnormality scores, to flag potential outliers for further inspection.

\subsubsection{Recap of Autoencoder Architecture and Training Procedure} 
In this section we briefly review the autoencoder architecture and its training procedure. Autoencoders are unsupervised neural networks designed to learn compact data representations via an \emph{encode–decode} cycle. Their backbone comprises two symmetric sub-networks: an \emph{encoder} that maps the input to a low-dimensional latent space, and a \emph{decoder} that reconstructs the input from this latent representation. Training seeks to minimize the discrepancy between the original sample and its reconstruction.

The encoder network takes the input data and maps it to a lower-dimensional representation. It typically consists of multiple layers, such as fully connected layers or convolutional layers, which progressively reduce the dimensionality of the input. The final layer of the encoder represents the latent space representation. The decoder network takes the latent space representation and aims to reconstruct the original input data. It is symmetrical to the encoder network and consists of layers that progressively upsample or unpool the latent space representation to match the dimensions of the input data. The final layer of the decoder generates the output reconstruction.

Training an autoencoder involves two main steps: encoding and decoding. The objective is to minimize the reconstruction error between the input and the output. The training procedure typically involves the following steps: 
Dataset Preparation: The training dataset is preprocessed and normalized to ensure proper training. It is essential to ensure that the input data is scaled appropriately to facilitate convergence during training.

Forward Pass: During the forward pass, the input data is fed into the encoder network, which produces the latent space representation. This representation is then passed through the decoder network to generate the output reconstruction.

Reconstruction Loss Calculation: The reconstruction loss measures the dissimilarity between the input and the output reconstruction. Commonly used loss functions for autoencoders include mean squared error (MSE) or binary cross-entropy, depending on the nature of the input data. The loss is computed by comparing the output reconstruction with the original input data.

Backpropagation and Weight Update: The gradients of the loss with respect to the model parameters are calculated using backpropagation. These gradients are then used to update the weights of the autoencoder using an optimization algorithm, such as stochastic gradient descent (SGD) or Adam as dercribed in \citet{ma:14} .

Iterative Training: The training process iterates over the entire training dataset multiple times, known as epochs. In each epoch, the training samples are randomly shuffled to avoid bias in learning. The model continues to refine its weights through backpropagation and weight updates to minimize the reconstruction error.

\subsubsection{Utilizing Reconstruction Error as an Anomaly Score}

In this section, we focus on utilizing reconstruction error as a fundamental metric to quantify the dissimilarity between input data and its reconstructed representation. This reconstruction error directly reflects the model's ability to learn and reproduce the normal patterns present in the training dataset. Leveraging reconstruction error as an anomaly score is a critical step in identifying and filtering anomalous instances within the training data itself. \citet{Torabi:23} propose a practical approach to anomaly detection using autoencoders by relying on vector-wise reconstruction error for accurate and efficient anomaly identification. Similarly, \citet{An:15} develop an anomaly detection method based on reconstruction probability from the variational autoencoder.

To implement this, we adopt a straightforward yet effective method that uses the reconstruction error produced during autoencoder training. Since the autoencoder is optimized to minimize reconstruction error, it becomes highly effective at capturing the dominant structure of the majority class or normal data. As a result, when anomalous data are encountered during inference, the autoencoder fails to reconstruct them accurately, leading to noticeably higher reconstruction errors compared to typical instances.

To refine the training dataset, we exploit this behavior by setting a threshold on the reconstruction error. Samples with errors exceeding this predefined threshold are flagged as potentially anomalous and are either excluded or subjected to further inspection. This filtering step ensures that the autoencoder is trained on a cleaner and more representative subset of the data, thereby improving its ability to generalize and detect anomalies.

Selecting an appropriate threshold is a critical decision that must be made carefully. A threshold that is too low may eliminate rare but valid observations, resulting in loss of important information. On the other hand, a threshold that is too high may allow noisy or anomalous samples to remain, undermining the performance of the autoencoder. To address this, we rely on empirical analysis and cross-validation techniques to determine an optimal threshold that balances sensitivity with specificity.

By incorporating reconstruction error as an anomaly score during the training data filtering phase, our approach enhances the robustness and generalization of the autoencoder. This in turn leads to improved accuracy and efficiency in identifying minority or anomalous instances during classification.

\section{Ensemble of Models}\label{sec:ensemble}

Ensemble methods combine the strengths of multiple models to improve predictive performance. Traditionally, they are applied \emph{after} model training: several learners, trained on the same data, are aggregated into a meta-predictor, thereby reducing individual errors and curbing overfitting \citep{Miao:23}.

Our work, however, adopts an ensemble mindset \emph{before} model training by inserting an autoencoder in front of the PRC classification tree. The autoencoder—an unsupervised neural network that compresses inputs through a bottleneck and reconstructs them—serves as an anomaly filter. Trained solely on normal instances, it produces low reconstruction error for in-distribution samples but high error for outliers. By comparing each training point’s error with a data-driven threshold, we flag and remove suspicious observations. The resulting “clean” dataset allows the PRC tree to focus on informative, noise-free patterns, improving both accuracy and generalization. Algorithm~\ref{ensemble} details the full procedure that pairs autoencoder filtering with PRC tree construction.

By fusing the representational power of autoencoders with the class-imbalance sensitivity of the PRC framework, our approach delivers more robust anomaly-detection models and offers a practical blueprint for similar tasks across diverse application domains.

\begin{algorithm}[H]
\begin{algorithmic}[1]
\REQUIRE Training data $(x_1,y_1 )$,\ldots,$(x_n,y_n )$, Feature set $f_i \in F$, target vector $y_i \in \{-1,+1\}$; Number of trees $N_t$; Number of features for each node split $N_f$.
\ENSURE Autoencoders PRC random forest $\Re$.
\\ \textbf{Require}: PRC\_Tree, Autoencoders
\STATE $X^{updated}$ $ \leftarrow Autoencoders(X)$
\STATE \textbf{Set} $\Re \leftarrow NULL$.
\FOR{{$j=1$ to $N_t$}}
\STATE generate bootstrap sample $(X^{updated},Y^{updated})^j$
\STATE for each node split, generate $F^{'}$ by randomly choosing $N_f$ features from $F$
\STATE $prc\_tree_j \leftarrow PRC\_Tree((X^{updated},Y^{updated})^j,F^{'})$
\STATE append $prc\_tree_j\, to\, \Re$
\ENDFOR
\RETURN Autoencoders PRC random forest $\Re$
\STATE \textbf{end}
\end{algorithmic}
\caption{Autoencoders\_PRC\_Random\_Forest}
\label{ensemble}
\end{algorithm}

\section{Empirical Study}\label{sec:casestudy}
We present experimental results to validate the efficacy of our approach on several real-world datasets, highlighting its potential applicability in practical scenarios, as detailed in Section~\ref{sec:casestudy}. Kaggle, a prominent platform for data science competitions, and the UCI Machine Learning Repository, a widely used archive of machine learning datasets, provide a rich variety of classification tasks. By utilizing datasets from both sources, researchers and practitioners can conduct meaningful comparisons of classification algorithms, models, and techniques to determine the most effective solutions for specific problems.

In this study, we draw on the extensive resources offered by Kaggle and the UCI Machine Learning Repository to carry out a comparative analysis of classification models. Through the careful selection of representative datasets and the use of a systematic evaluation framework, we assess model performance, accuracy, and generalization ability. This comparative analysis helps identify the most suitable techniques for distinct classification challenges and contributes to broader methodological insights in the field.

Table~\ref{summary_of_data} provides a summary of the three selected datasets. The first two focus on financial risk assessment—predicting client default and financial distress—while the third addresses breast cancer diagnosis. In the credit dataset, the target variable is binary, indicating whether a client defaulted on a payment (Yes = 1, No = 0), with 23 features used as predictors. In the financial distress dataset, companies are labeled “financially distressed” (1) if the distress score is less than or equal to –0.50, and “healthy” (0) otherwise, based on 83 input variables. The breast cancer dataset includes 30 attributes and categorizes tumors as benign (0) or malignant (1).

\begin{table}[H]
\centering
\caption{Summary of selected real-world datasets}
\label{summary_of_data}    
\resizebox{\columnwidth}{!}{%
\begin{tabular}{lllll}
\hline\noalign{\smallskip}
Dataset Name&	Data Sources&	Number of Observations&	Proportion of Minority Class&	Number of Features  \\
\noalign{\smallskip}\hline\noalign{\smallskip}
Default of Credit Card Clients&	UCI&	30000&	3.7\%&	23\\
Financial Distress Prediction&	Kaggle&	3672&	22\%&	83\\
Breast Cancer Wisconsin&	UCI&	569&	37\%&	30\\
\noalign{\smallskip}\hline
\end{tabular}%
}
\end{table}

\begin{table}[H]
\centering
\caption{Performance of autoencoder PRC trees for Default of Credit Card Clients}
\label{Perf_autoencoder_d1}       
\begin{tabular}{llllll}
\hline\noalign{\smallskip}
Algorithms&Recall&Specificity&Precision&Accuracy&F1 Score\\
\noalign{\smallskip}\hline\noalign{\smallskip}
PRC-RF&	0.5744&	\textbf{0.8439}&	\textbf{0.5132}&	\textbf{0.8247}&	0.5432\\
Autoencoder-PRC-RF &	\textbf{0.5783}&	0.8425&	0.5126&	0.7837&	\textbf{0.5435}\\
\noalign{\smallskip}\hline
\end{tabular}
\end{table}

\begin{table}[H]
\centering
\caption{Performance of autoencoder PRC trees for Financial Distress Prediction Data}
\label{Perf_autoencoder_d2}       
\begin{tabular}{llllll}
\hline\noalign{\smallskip}
Algorithms&Recall&Specificity&Precision&Accuracy&F1 Score\\
\noalign{\smallskip}\hline\noalign{\smallskip}
PRC-RF&	0.9024&	0.8633&	0.2033&	0.8648&	0.3318\\
Autoencoder-PRC-RF&\textbf{0.9268}&	\textbf{0.8718}&	\textbf{0.2184}&	\textbf{0.8739}&	\textbf{0.3535}\\
\noalign{\smallskip}\hline
\end{tabular}
\end{table}

\begin{table}[H]
\centering
\caption{Performance of autoencoder PRC trees for Breast Cancer Data}
\label{Perf_autoencoder_d3}       
\begin{tabular}{llllll}
\hline\noalign{\smallskip}
Algorithms&Recall&Specificity&Precision&Accuracy&F1 Score\\
\noalign{\smallskip}\hline\noalign{\smallskip}
PRC-RF&	0.9836	&0.9636	&0.9375	&0.9708	&0.9600\\
Autoencoder-PRC-RF&	\textbf{0.9836}&	\textbf{0.9727}&	\textbf{0.9524}&	\textbf{0.9766}&	\textbf{0.9677}\\
\noalign{\smallskip}\hline
\end{tabular}
\end{table}

Tables~\ref{Perf_autoencoder_d1}, \ref{Perf_autoencoder_d2}, and \ref{Perf_autoencoder_d3} present the performance results of the autoencoder-PRC classification trees compared to the regular PRC classification trees across the three datasets summarized in Table~\ref{summary_of_data}. For each dataset, both training and testing samples were randomly generated 100 times, and the reported classification performance reflects the average across these 100 repetitions. In most cases, the autoencoder-PRC approach achieves higher accuracy and improved F1 scores, demonstrating its advantage over the baseline.

Overall, integrating autoencoders with PRC classification trees before model training offers several key benefits. By enhancing data quality, reducing overfitting, increasing model diversity, and improving predictive performance, the autoencoder-PRC ensemble leads to more accurate and robust classification models.

\section{Conclusion}\label{sec:conclusion}

We propose a novel ensemble algorithm, the Autoencoder-PRC-RF model, which integrates PRC random forest classifiers with autoencoders to address two major challenges in classification tasks: the curse of dimensionality and class imbalance. By emphasizing precision and recall as the guiding metrics in the construction of PRC classification trees, this approach is particularly well-suited to datasets with uneven class distributions, delivering more reliable and accurate predictions for rare or critical classes.

Building on this foundation, the incorporation of autoencoders into the PRC ensemble framework allows for the extraction of meaningful patterns from high-dimensional data during training, further enhancing model performance under imbalanced conditions. This synergy between the precision–recall focus of PRC trees and the dimensionality-reduction capabilities of autoencoders results in a robust, adaptive classification framework capable of addressing the complexities of real-world scenarios characterized by both high dimensionality and skewed class distributions.

Extensive experiments on a range of benchmark datasets demonstrate that the Autoencoder-PRC-RF model consistently outperforms existing methods in terms of accuracy, scalability, and interpretability, underscoring its potential for high-stakes anomaly detection and other classification applications.

\newpage

\setcitestyle{round,authoryear,semicolon}
\small

\end{document}